\documentclass[10pt,twocolumn,letterpaper]{article}

\usepackage{iccv}
\usepackage[dvipdfmx]{graphicx}
\usepackage{times}
\usepackage{epsfig}
\usepackage{graphicx}
\usepackage{amsmath}
\usepackage{amssymb}
\usepackage{here}
\usepackage{url}
\usepackage{multicol}
\DeclareMathOperator*{\argmax}{arg\,max}

\usepackage[breaklinks=true,bookmarks=false]{hyperref}

\iccvfinalcopy 

\ificcvfinal\fi
\begin{document}

\title{Generating Easy-to-Understand Referring Expressions for Target Identifications}

\author{
    Mikihiro Tanaka${}^\text{1}$, Takayuki Itamochi${}^\text{2}$, Kenichi Narioka${}^\text{2}$,
    Ikuro Sato${}^\text{3}$,\\ Yoshitaka Ushiku${}^\text{1}$ and Tatsuya Harada${}^\text{1,4}$\\
    ${}^\text{1}$The University of Tokyo, ${}^\text{2}$DENSO CORPORATION, ${}^\text{3}$DENSO IT Laboratory, Inc., ${}^\text{4}$RIKEN
}

\maketitle
\ificcvfinal\thispagestyle{empty}\fi

\begin{abstract}
This paper addresses the generation of referring expressions that not only refer to objects correctly but also let humans find them quickly. As a target becomes relatively less salient, identifying referred objects itself becomes more difficult. However, the existing studies regarded all sentences that refer to objects correctly as equally good, ignoring whether they are easily understood by humans. If the target is not salient, humans utilize relationships with the salient contexts around it to help listeners to comprehend it better. To derive this information from human annotations, our model is designed to extract information from the target and from the environment. Moreover, we regard that sentences that are easily understood are those that are comprehended correctly and quickly by humans. We optimized this by using the time required to locate the referred objects by humans and their accuracies.
To evaluate our system, we created a new referring expression dataset whose images were acquired from Grand Theft Auto V (GTA V), limiting targets to persons. Experimental results show the effectiveness of our approach. Our code and dataset are available at https://github.com/mikittt/easy-to-understand-REG.
\end{abstract}
\vspace{-11pt}

\section{Introduction}
\vspace{-3pt}

\begin{figure}[t]
\vspace{-6pt}
\begin{center}
   \includegraphics[width=0.93\linewidth]{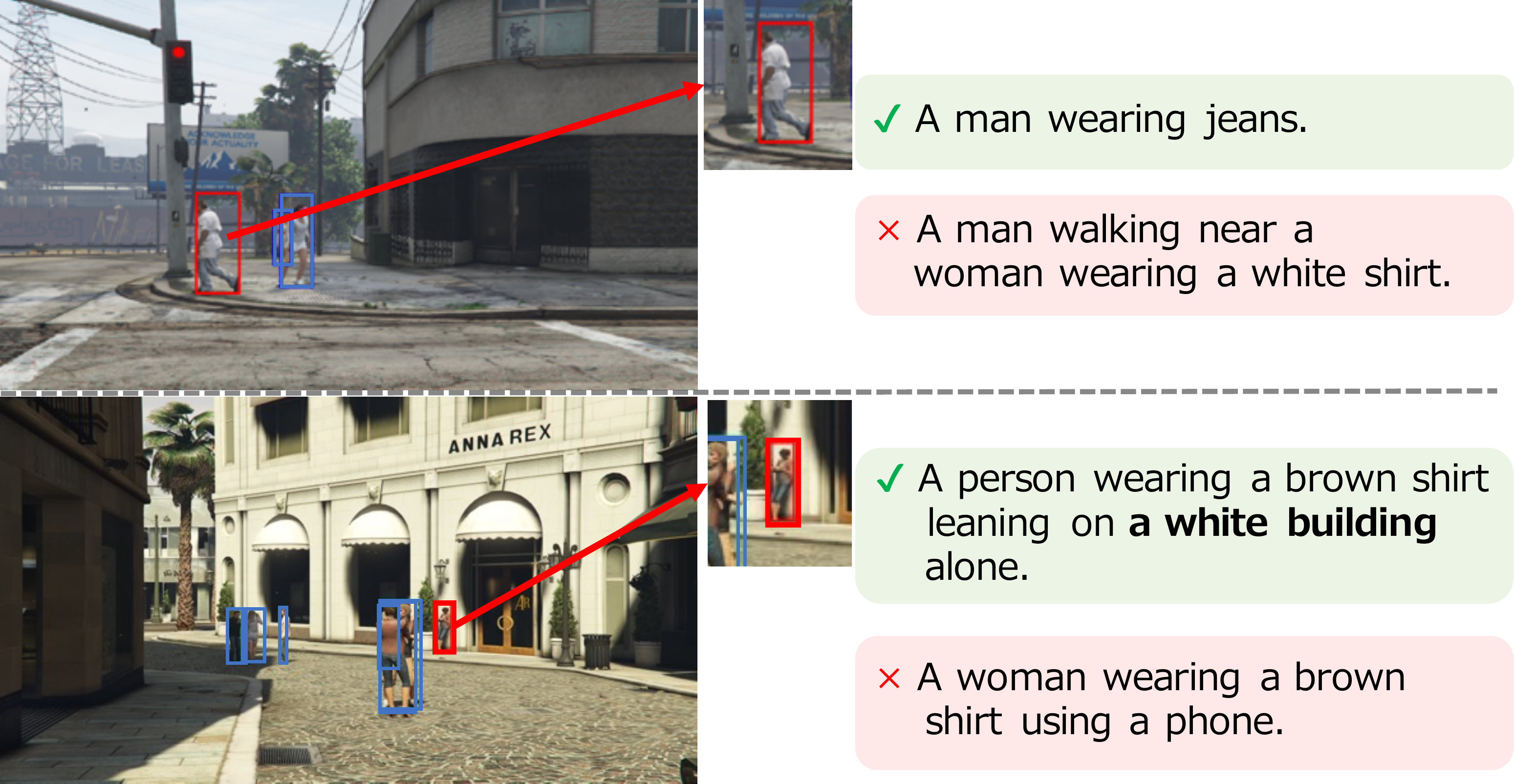}
\end{center}
\vspace{-10pt}
   \caption{Examples of referring expressions to be generated in this study. In the top image, the target in the red bounding box is sufficiently salient; therefore, a brief description suffices. In the bottom image, referring to other salient objects is required to single out the target because the target itself is not sufficiently salient.}
\vspace{-8pt}
\label{fig:theme}
\label{fig:long}
\label{fig:onecol}
\end{figure}

With the popularization of intelligent agents such as robots, symbiosis with them becomes more important. Sharing what humans and agents see naturally is a particularly essential component for smooth communication in the symbiosis environment. In daily life, people often use referring expressions to indicate specific targets such as ``a man wearing a red shirt.'' Further, communicating with agents with natural language is an intuitive method of interaction. When referring to a object with natural language, many expressions can be used that are equally \textit{correct} from a semantic standpoint such that one can locate the target. However, they are not always equally \textit{easy} for target identifications. As shown in Fig.~\ref{fig:theme}, it is important for the expression to be comprehended easily by humans. The comprehension is divided into two processes: understanding the text and finding an referred object in an image. These can be uniformly evaluated by the comprehension time by humans. Thus, we regard that easy-to-understand referring expressions are those that are comprehended correctly and quickly by humans.

Recently, \textit{correct} referring expression generation has demonstrated significant progress. Considering agents' views that are automatically captured such as in-vehicle images, the compositions of the images are often complex and contain more objects with low saliency than images from MSCOCO~\cite{Lin2014}, which are typically used in the existing works of referring expression generation~\cite{Liu2017,Mao2016,Yu2016,Yu2017,Luo2017}. The existing studies regarded expressions that refer to objects correctly as equally good. However, if the targets become relatively less salient, identifying the referred objects can become difficult even if the sentences are \textit{correct}.

For the agents to refer to objects in natural language, they should be described clearly for an \textit{easier} comprehension. Expressions utilizing relationships between the targets and other salient contexts such as ``a woman by the red car'' would help listeners to identify the referred objects when the targets are not sufficiently salient. Thus, expressions to be generated demand the following properties:
\begin{itemize}
 \setlength{\parskip}{0cm} 
 \setlength{\itemsep}{0cm} 
\vspace{-8pt}
\item If the target is salient, a brief description suffices.
\item If the target is less salient, utilizing relationships with salient contexts around it helps to tell its location.
\vspace{-8pt}
\end{itemize}
If these sentences can be generated, drivers can be navigated by utilizing in-vehicle images such as, ``please turn right at the intersection by which a man with a red cap stands.''

We herein propose a new approach to generate referring expressions that are brief and sufficiently easy for a human to locate a target object in an image without sacrificing the semantic validity. To utilize salient contexts around the target, our model is designed to extract information from the target and from the environment. Moreover, we perform optimization for the expression generator using the time and accuracy metrics for target identifications. Although these quantities by themselves do not tell the absolute level of the goodness of the generated sentences, comparing them among candidate sentences helps to identify a preferable one. We adopt a ranking learning technique in this respect.


To evaluate our system, we constructed a new referring expression dataset with images from GTA V~\cite{GTAV}, limiting targets to humans for two reasons. (1) Targeting humans is of primary importance for the symbiosis of humans and robots as well as in designing safe mobile agents. The existence of pedestrian detection field~\cite{Oren1997,Papageorgiou2000,Dollar2009,Zhang2016,Brazil2017,Zhang_2018} also tells the importance of application. (2) Targeting humans is technically challenging because humans have various contexts as they act in various places and their appearances vary widely. We included humans' comprehension time and accuracy in the dataset for the ranking method above.


Overall, our primary contributions are as follows.
\begin{itemize}
 \setlength{\parskip}{0cm} 
 \setlength{\itemsep}{0cm} 
\vspace{-7pt}
 \item We propose a novel task whose goal is to generate referring expressions that can be comprehended correctly and quickly by humans. 
 \item We propose a optimization method for the task above with additional human annotations and a novel referring expression generation model which captures contexts around the targets.
 \item We created a new large-scale dataset for the task above based on GTA V (RefGTA), which contains images with complex compositions and more targets with low saliency than the existing referring expression datasets.
 \item Experimental results on RefGTA show the effectiveness of our approach, whilst the results on existing datasets show the versatility of our method on various objects with real images.
\vspace{-5pt}
\end{itemize}

\section{Related work}
\vspace{-2pt}
First, we introduce image captioning. Next, we explain referring expression generation that describes specific objects. Finally, we refer to datasets used for referring expression generation and comprehension.

\vspace{-2pt}
\subsection{Image Captioning}
\vspace{-2pt}
Following the advent in image recognition and machine translation with deep neural networks, the encoder-decoder model improved the quality of image captioning significantly, which encodes an image with a deep convolutional neural network (CNN), and subsequently decodes it by a long term-short memory (LSTM)~\cite{Vinyals2015}. Many recent approaches use attention models that extract local image features dynamically while generating each word of a sentence~\cite{Lu2017, Xu2015, Rennie2017, Li2017, Anderson2018, Lu2018, Yao2018}. Lu \etal~\cite{Lu2017} introduced a new hidden state of the LSTM called the ``visual sentinel'' vector. It controls when to attend the image by holding the context of previously generated words, because words such as ``the'' and ``of'' depend on the sentence context rather than the image information. Recently, researchers have applied reinforcement learning to directly optimize automatic evaluation metrics that are non-differentiable~\cite{Rennie2017, Anderson2018, Lu2018, Yao2018}.

\vspace{-2pt}
\subsection{Referring Expression Generation}
\vspace{-2pt}
While image captioning describes a full image, referring expression generation is to generate a sentence that distinguishes a specific object from others in an image. Referring expressions have been studied for a long time as a NLP problem~\cite{Winograd1972,Krahmer2012}. Recently, large-scale datasets (RefCOCO, RefCOCO+~\cite{Yu2016}, RefCOCOg~\cite{Mao2016}, etc.) were constructed, and both referring expression generation and comprehension have been developed in pictures acquired in the real world~\cite{Mao2016,Yu2016,Yu2017,Nagaraja16,Liu2017,Luo2017,Hu2017,Yu2018,Zhang2018,Vasudevan2018}. As these problems are complementary,
recent approaches of referring expression generation solve both problems simultaneously~\cite{Liu2017,Mao2016,Yu2016,Yu2017,Luo2017}. Mao \etal~\cite{Mao2016} introduced max-margin Maximum Mutual Information (MMI) training that solves comprehension problems with a single model to generate disambiguous sentences. Liu \etal~\cite{Liu2017} focused on the attributes of the targets and improved the performance. Yu \etal~\cite{Yu2017} proposed a method that jointly optimizes the speaker, listener, and reinforcer models, and acquired state-of-the-art performance. Their respective roles are to generate referring expressions, comprehend the referred objects, and reward the speaker for generating discriminative expressions. 

\vspace{-2pt}
\subsection{Referring Expression Datasets}
\vspace{-2pt}
The initial datasets consist of simple computer graphics~\cite{Viethen2008} or small natural objects~\cite{Mitchell2010, FitzGerald2013}. Subsequently, first large-scale referring expression dataset RefCLEF~\cite{Kazemzadeh2014} was constructed using images from ImageClef~\cite{Escalante2010}. By utilizing images from MSCOCO, other large-scale datasets such as RefCOCO, RefCOCO+ and RefCOCOg were collected. These useful datasets consist of many images captured by humans, whose compositions are simple with some subjects in the center. For images captured by robots or other intelligent agents, handling more complex images is important. Some existing studies constructed referring expression datasets with images~\cite{Zhou2017} or videos~\cite{Vasudevan2018} from Cityscapes~\cite{Cordts2016}. However, this is created for comprehension and the sentence should just refer to the target correctly because the listeners are supposed to be machines. We focus on generation and the understandability of the sentence should be considered because the listeners are supposed to be humans. In this respect, we created a new dataset with images from GTA V described in Sec.~\ref{sec:dataset_construction}.

\begin{figure*}[t!]
\vspace{-9pt}
\centering
  \includegraphics[width=\linewidth]{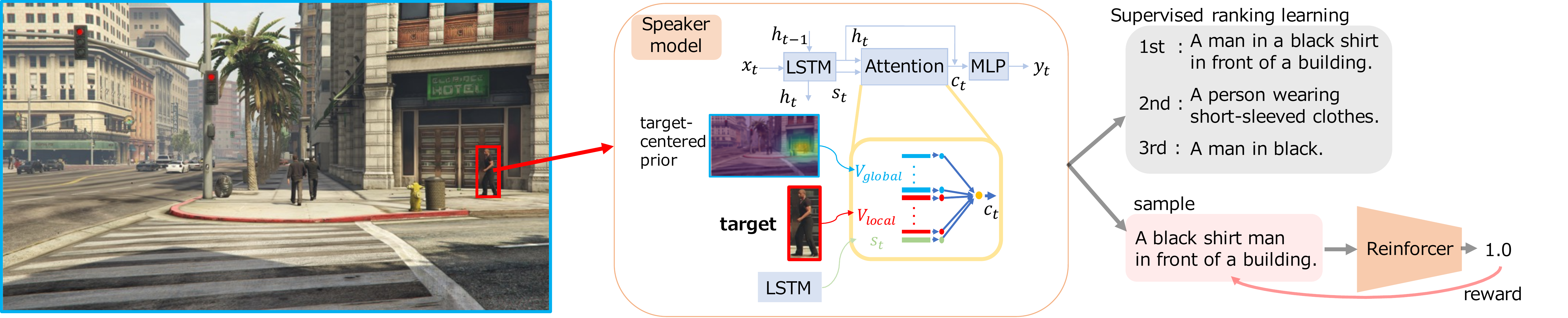}
  \caption{Our model consists of two components. The first one \textit{speaker} is in the middle of the figure. \textit{Speaker} is trained to generate referring expressions with supervised ranking learning and the reward from the second model \textit{reinforcer} in the right side of the figure. \textit{Speaker} attends to features from the target, the context around it, and the sentence context under generation $s_t$.}
\label{fig:arch}
\vspace{-12pt}
\end{figure*}

\section{Model}
\vspace{-2pt}
\label{sec:model}
To generate easy-to-understand referring expressions for target identifications, the model should be able to inform us of the target's location utilizing salient context around it. Similar to normal image captioning, we consider generating sentences word by word, and the context of the sentence information under generation is also utilized. We refer to this context as the sentence context. We assumed the necessary information to generate the sentences as follows.
\begin{itemize}
 \setlength{\parskip}{0cm} 
 \setlength{\itemsep}{0cm} 
 \vspace{-6pt}
 \item[(A)] Salient features of the target
 \item[(B)] Relationships between the target and salient context around it
 \item[(C)] Sentence context under generation
 \vspace{-6pt}
\end{itemize}

We propose a model comprising a novel context-aware
speaker and reinforcer. For the context novelty, please see our supplementary material. As reported in~\cite{Yu2017}, joint optimization using both a listener and reinforcer achieves similar performance to using either one in isolation. This is mainly because both of them provide feedback to the neural network based on the same ground truth captions. Instead, we aim to generate more appropriate captions by modifying the speaker given the above assumptions (A), (B) and (C).

Moreover, expressions to be generated should help a human in locating the referred objects correctly and quickly. If the targets are sufficiently salient, brief expressions are preferable for rapid comprehension. We optimized them by comparing the time required to locate the referred objects by humans, and their accuracies among sentences annotated to the same instance. 

First, we introduce a state-of-the-art method to generate referring expressions, i.e., the speaker-listener-reinforcer~\cite{Yu2017}. Next, we explain our generation model. Finally, we introduce the optimization of easy-to-understand referring expressions and describe compound loss.

\vspace{-2pt}
\subsection{Baseline Method}
\vspace{-2pt}
We explain a state-of-the-art method~\cite{Yu2017}. Three models, speaker, listener, and reinforcer were used.
Herein, we explain only the speaker and reinforcer that are used in our proposed model.

\noindent{\textbf{Speaker:}} For generating referring expressions, the speaker model should extract target object features that are distinguished from other objects. Yu \etal~\cite{Yu2017} used the CNN to extract image features and generate sentences by LSTM. First, Yu \etal~\cite{Yu2017} extracted the following five features:
(1) target object feature vector $o_i$, (2) whole image feature vector ${g}_i$, (3) the feature encoding the target's coordinate $(x,y)$ and the size $(w, h)$ as $l_i=[\frac{x_{tl}}{W},\frac{y_{tl}}{H},\frac{x_{br}}{W},\frac{y_{br}}{H},\frac{w_i \cdot h_i}{W \cdot H}]$, (4) difference in target object feature from others $\delta o_i=\frac{1}{n}\sum_{j\not= i}\frac{o_i-o_j}{\|o_i-o_j\|}$, (5) difference in target coordinate from others $\delta l_{ij}=[\frac{[\Delta x_{tl}]_{ij}}{w_{i}},\frac{[\Delta y_{tl}]_{ij}}{h_{i}},\frac{[\Delta x_{br}]_{ij}}{w_{i}},\frac{[\Delta y_{br}]_{ij}}{h_{i}},\frac{w_j\cdot h_j}{w_i\cdot h_i}]$. Visual feature $v_i$ is obtained by applying one linear layer to these, $v_i=W_{m}[o_i,{g}_i,l_i,\delta o_i,\delta l_{i}]$. Concatenating $v_i$ and the word embedding vector $w_t$, $x_t=[v_i;w_t]$ is fed into the LSTM and learned to generate sentences $r_i$ by minimizing the negative log-likelihood with model parameters $\theta$:
\vspace{-1pt}
\begin{equation}
L^1_s(\theta) = -\sum_i \log P(r_i| v_i;\theta)
\label{eqn:log_likeli}
\end{equation}
\vspace{-6pt}

To generate discriminative sentences, they generalized the MMI~\cite{Mao2016} to enforce the model, to increase the probability of generating sentences $r_i$ if the given positive pairs as ($r_i$,$o_i$) than if the given negative pairs as ($r_j$,$o_i$) or ($r_i$,$o_k$) where $r_j$ and $o_k$ are sampled randomly from other objects, and optimized by following the max-margin loss ($\lambda^s_1$, $\lambda^s_2$, $M_1$ and $M_2$ are hyper-parameters):
\vspace{-2pt}
\begin{equation}
\scalebox{0.88}{$\displaystyle
\begin{split}
L^2_s(\theta) =\sum_{i} \{&\lambda^s_1 \max(0, M_1+\log P(r_i|v_k)-\log P(r_i|v_i))\\
+&\lambda^s_2 \max(0, M_2+\log P(r_j|v_i)-\log P(r_i|v_i))\}
\label{eqn:mmi}
\end{split}$}
\end{equation}

\noindent{\textbf{Reinforcer:}} Next, we explain the reinforcer module that rewards the speaker model for generating discriminative sentences. First, the reinforcer model is pretrained by classifying whether the input image feature and sentence feature are paired by logistic regression. The reinforcer extracts image features by the CNN and sentence features by LSTM and subsequently, feed into MLP by concatenating both features to output a scalar. Next, it rewards the speaker model while fixing its parameters. Because the sampling operation of sentences $w_{1:T}$ is non-differentiable, they used policy-gradient to train the speaker to maximize the reward $F(w_{1:T}, o_i)$ by the following loss:
\vspace{-2pt}
\begin{equation}
\nabla_{\theta}J = 
-E_{P(w_{1:T}|v_i)}[(F(w_{1:T}, v_i))\nabla_{\theta}\log P(w_{1:T}|v_i;\theta)]
\label{eqn:policy}
\end{equation}

\subsection{Context-aware speaker model}
Our speaker model (in Fig.~\ref{fig:arch}) generates referring expressions that can utilize relationships between targets and salient contexts around the target. Similar to Yu~\cite{Yu2017}, we encoded image features by the CNN, and decoded it into a language by LSTM. In extracting the global features from whole images whose compositions are complex, information around the target objects is more important. We replace global features $g_i$ with ${g}^{\prime}_i$ that weight Gaussian distribution whose center is the center of the target (variance is a learnable parameter). We used $v_i=W_{m}[o_i,{g}^{\prime}_i,l_i,\delta o_i,\delta l_{ij}]$ as a target image feature to feed into the LSTM.

Next, we introduce the attention module that satisfies the requirements. We begin by defining the notations: $V_{{\rm global}}$, $V_{{\rm local}}$ are the output features of the last convolutional layer on the CNN, containing $k, l$ spatial features respectively ($V_{\rm global} = [{f^g}_1, \cdots , {f^g}_k] , V_{{\rm local}} = [{f^l}_1, \cdots , {f^l}_l]$, $V_{{\rm global}}\in\mathbb{R}^{d \times k}, V_{{\rm local}}\in\mathbb{R}^{d \times l}$).
To extract the required information: (A) salient features of the target, (B) relationships with salient context around it, and (C) sentence context under generation, we can use $V_{\rm local}$, $V_{\rm global}$ for (A) and (B), respectively. As for (C), we used a sentinel vector $s_t$ proposed by Lu \etal~\cite{Lu2017}, which is a hidden state of the LSTM calculated as follows: ($h_t$: hidden state of LSTM, $m_t$: memory cell of LSTM):
\vspace{-2pt}
\begin{eqnarray}
&s_t = \sigma (W_x x_t + W_h h_{t-1}) \odot \tanh{(m_t)}
\end{eqnarray}
 For focusing more around the target, we introduce target-centered weighting $G_i$ ($G_i\in\mathbb{R}^{1 \times k}$) with Gaussian distribution, similar as in the feature ${g}^{\prime}_i$. Using four weights, $W_{{\rm global}}\in\mathbb{R}^{d \times d},W_{{\rm local}}\in\mathbb{R}^{d \times d},W_{s}\in\mathbb{R}^{d \times d},w_{h}\in\mathbb{R}^{d \times 1}$, and defining $V_t=[V_{\rm global};V_{\rm local};s_t]$, our attention $\alpha_t$ is calculated as follows:
\vspace{-2pt}
\begin{flalign}
v_t &= [W_{{\rm global}}V_{{\rm global}} ; W_{{\rm local}}V_{{\rm local}_i} ; W_{s}s_t]\\
z_t &= w_h^T \tanh(v_t+W_{g}h_{t}1^T)\\
\alpha_t &= {\it {\rm Softmax}}([(z_t[:,:k]+\log{G_i}) ; z_t[:,k:]])
\label{eqn:attn}
\end{flalign}
([;] implies concatenation, and [:,:$k$] implies to 
extract the partial matrix up to column $k$)

Finally, we can obtain the probability of possible words as follows:
\vspace{-10pt}
\begin{flalign}
c_t &= \sum_{n=1}^{k+l+1} \alpha_{tn}V_{tn}\\
p(w_t|w_1, \cdots ,w_{t-1}, v_i) &= {\rm Softmax}(W_p(c_t+h_t))
\end{flalign} 

\subsection{Optimization of easy-to-understand referring expressions\label{sec:ranking_opt}}
In our task, sentences to be generated should be comprehended by humans (1) correctly and (2) quickly. Although (1) can be learned by the baseline method, (2) is difficult to optimize because defining an absolute indicator that can measure it is difficult. However, we can determine which sentence is better than the others by human annotations. In our task, we used the time required by humans to identify the referred objects and its accuracy for the annotations.

We now consider ranking labels as teacher information. For a target $o_i$, sentences $\{r_{i1},\cdots,r_{im}\}$ are annotated. We denote a set of pairs satisfying rank($r_{ip}$) $<$ rank($r_{iq}$) ($p\neq q, 1 \leq p,q\leq m$) as $\Omega_i$. In this case, the probability of generating $r_{ip}$ should be higher than one of generating $r_{iq}$. We sample ($r_{ip}$, $r_{iq}$) randomly from $\Omega_i$ and perform optimization by the max margin loss as follows ($\lambda^s_3$ and $M_3$ are hyper-parameters):
\begin{equation}
\label{eqn:rank_los}
\scalebox{0.9}{$\displaystyle
L^3_s(\theta) = \sum_i \{\lambda^s_3 \max(0, M_3+\log P(r_{iq}|v_i)-\log P(r_{ip}|v_i))\}
$}
\end{equation}

Moreover, we applied this ranking loss to the reinforcer model. We used the output before the last sigmoid activation to calculate the loss similar to the above Eqn.~\ref{eqn:rank_los}. The final loss function of the reinforcer is both the ranking loss and logistic regression. Similar to  Eqn.~\ref{eqn:policy}, we can train the speaker to generate sentences to maximize the new reward $F^{\prime}(w_{1:T}, o_i)$, which estimates how easily the generated expressions can be comprehended by humans as follows:
\begin{equation}
\nabla_{\theta}J^{\prime} = 
-E_{P(w_{1:T}|v_i)}[(F^{\prime}(w_{1:T}, v_i))\nabla_{\theta}\log P(w_{1:T}|v_i;\theta)]
\label{eqn:policy2}
\end{equation}

We also introduced sentence attention~\cite{Yuvqa2017} into the sentence encoder of the model to capture the words that would facilitate a human's comprehension of a sentence.

\noindent{\textbf{Compund loss:}} The final loss of our speaker model $L_s$ is a combination of Eqn.~\ref{eqn:log_likeli}, Eqn.~\ref{eqn:mmi}, Eqn.~\ref{eqn:rank_los} and Eqn.~\ref{eqn:policy2} as follows ($\lambda^r$ is a hyper-parameter.):
\begin{flalign}
&L_s(\theta) = L^1_s+L^2_s+L^3_s-\lambda^r J^{\prime}
\end{flalign}

\vspace{-2pt}
\section{Dataset Construction}
\vspace{-2pt}
\label{sec:dataset_construction}
\begin{figure*}[tbp]
\vspace{-6pt}
\centering
  \includegraphics[width=0.94\linewidth]{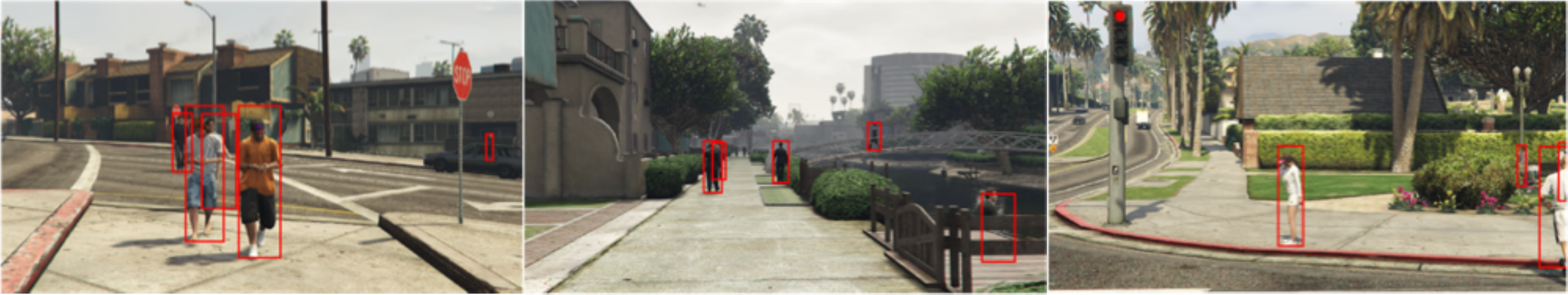}
  \vspace{-2pt}
  \caption{Images from GTA V. Left : images with unconstrained clothing / Middle : images in which only black-clothed persons exist / Right : images in which only white-clothed persons exist}
\label{fig:image_collect}
\vspace{-16pt}
\end{figure*}

In our task, the following properties are required in the dataset. (1) The composition of the images are complex. (2) Targets' appearances and locations are sufficiently diverse. However, dataset bias as for (2) tends to occur when collecting a real dataset. We acquired images which satisfy (1) from GTA V because CG can be easily controlled and can guarantee (2). Artificial datasets such as CLEVR~\cite{Johnson2017} are also advantageous as they can isolate and control the qualitative difficulty of the problem and are widely applied in similar problem settings. For real world applications, synthetic data can help improve understanding as in~\cite{Varol2017} and we can also use unsupervised domain adaptation as in~\cite{Hoffman2018}. In this study we constructed a new referring expression dataset, RefGTA, limiting the target type to humans only.

We collected images and information such as a person's bounding boxes automatically, and subsequently annotated the referring expression by humans. (GTA V is allowed for use in non-commercial and research uses~\cite{GTAPolicy}.)

\subsection{Image Collection}
\label{sec:image_collect_section}
\vspace{-4pt}
First, we extracted images and persons' bounding box information once every few seconds using a GTA V mod that we created (PC single-player mods are allowed~\cite{GTAMod}).

Moreover, even when multiple persons whose appearances are similar exist, the system should be able to generate expressions where humans can identify referred objects easily by utilizing the relationships between the targets and other objects \etc. Therefore, we further collected images in which only either white-clothed or black-clothed persons exist, by setting them when the mod starts, as in Fig.~\ref{fig:image_collect}.

Finally, we deleted similar images by the average hash. We also deleted images comprising combinations of the same characters. We set the obtained images and bounding box information as a dataset.

\vspace{-4pt}
\subsection{Sentence Annotation}
\label{sec:sent_ann}
\vspace{-4pt}
We annotate sentences to each instance obtained in Sec.~\ref{sec:image_collect_section} by the following two steps. We requested the annotations of the Amazon Mechanical
Turk (AMT) workers.

\noindent{\textbf{Annotating sentences:}} First, we requested the AMT workers to annotate five descriptions that are distinguished from the others for each instance. We instructed the workers to annotate a sentence that refers only to the target and is easy to distinguish from others at a glance. We also instructed the workers to use not only the target attributes but also the relative positions to other objects in the image. We instructed them not to use absolute positions inside the image and allowed the relative positions to other objects.

\noindent{\textbf{Validating sentences:}} Next, we assigned five AMT workers to localize the referred person in each description to verify whether it is an appropriate referring expression. If a referred person does not exist, we allow them to check the box, ``impossible to identify.'' We displayed the elapsed time on the task screen and instructed the workers to obtain the referred objects as quickly as possible. We included the sentences where more than half of the workers accurately obtained the referred persons in a dataset. We also recorded the time and accuracy of five workers for each sentence.

\begin{figure}[tbp]
\vspace{-11pt}
\begin{center}
  \includegraphics[width=\linewidth]{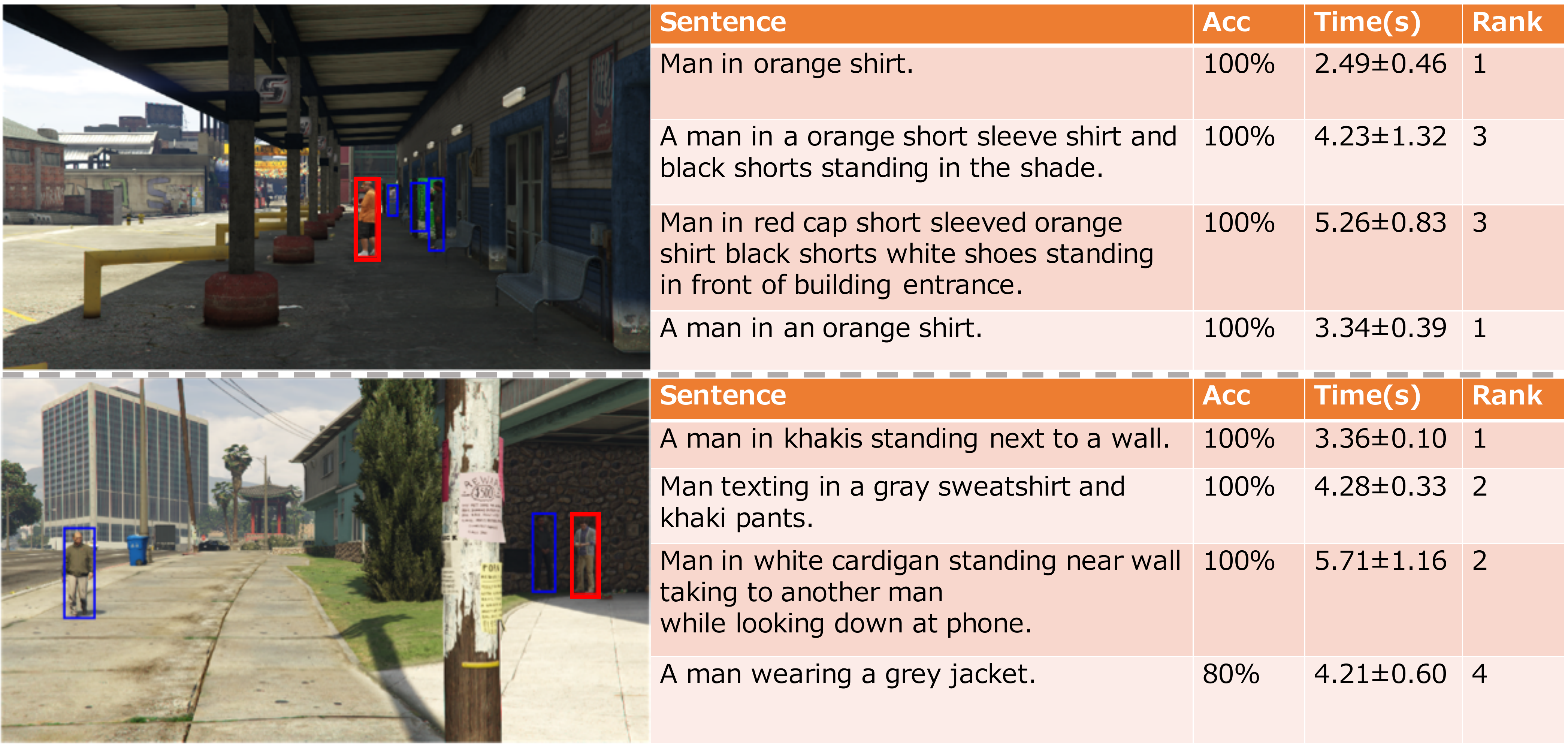}
  \caption{Example data. Sentence: Annotated captions. Acc: Human's comprehension accuracy. Time (s): the time required by human to search. Rank: The ranking we assigned by the accuracy and time as described in this section.}
\label{fig:annotation1}
\vspace{-10pt}
\end{center}
\end{figure}

\noindent{\textbf{Examples:}} We show the annotation examples in Fig.~\ref{fig:annotation1}. The rightmost column is the ranking we used in Sec.~\ref{sec:ranking_opt}. This is calculated as follows: first, all sentences are ranked by humans' comprehension accuracy; subsequently, sentences that are comprehended correctly by all workers are ranked by time. This ranking is performed as follows. When comparing the times of two sentences we take the time of three people in the middle of five people to reduce the influence of outliers. We consider sentence ``A'' as better than sentence ``B'' if the mean of ``B'' subtracted by the mean of ``A'' is greater than the sum of their standard errors. For each sentence we count the number of sentences that it is better than and rank the sentence according to this number.

\vspace{-2pt}
\subsection{Statistical Information}
\label{sec:stat_info} 
\vspace{-3pt}
\begin{table}[tbp]
\vspace{-8pt}
\begin{center}
\resizebox{0.65\columnwidth}{!}{%
\begin{tabular}{|l|cc|c|c|}
\hline
 & Train & Val & Test \\
\hline\hline
\# of images  & 23,950 & 2,395 & 2,405 \\
\# of created instances & 65,205 & 6,563 & 6,504 \\
\# of referring expressions & 177,763 & 17,766 & 17,646 \\
\hline
\end{tabular}}
\caption{Statistics of annotations on the dataset.}
\label{table:annotation_volumes}
\end{center}
\vspace{-11pt}
\end{table}

\begin{figure}[tbp]
\centering
\vspace{-8pt}
  \includegraphics[width=0.75\linewidth]{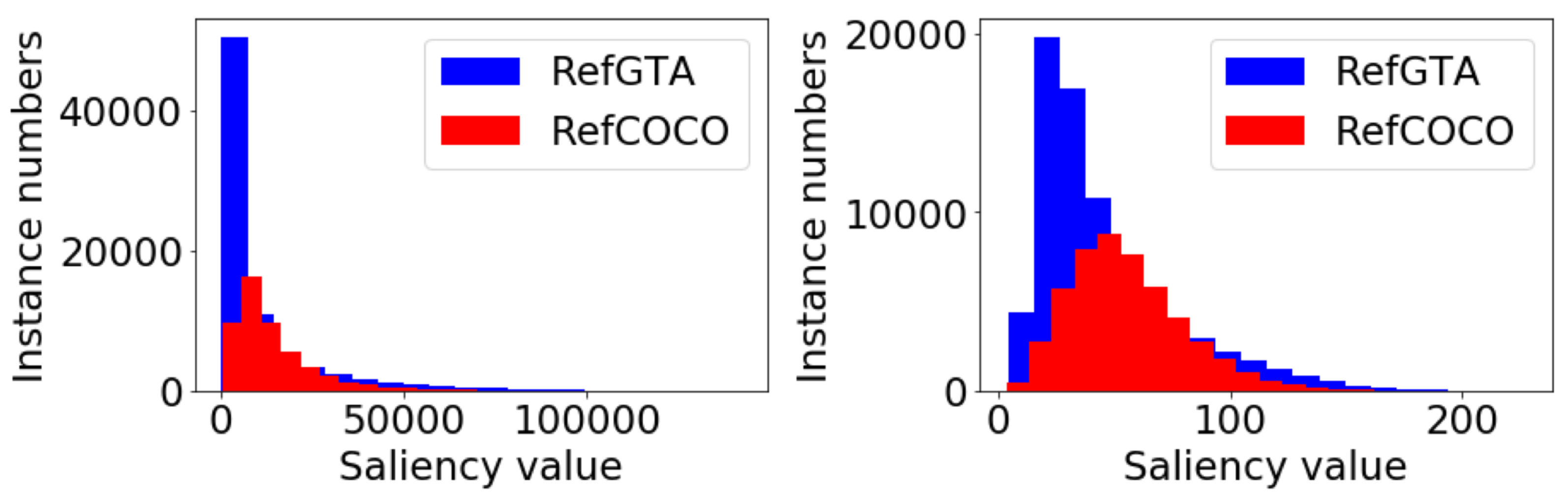}
  \vspace{-3pt}
  \caption{Targets' saliency of RefCOCO and RefGTA. Left: saliency is calculated by the sum of the saliency score inside the target bounding box. Right: 
saliency is normalized by dividing the square root of the area.}
\label{fig:salient_compare}
\vspace{-13pt}
\end{figure}

We show the statistics of our dataset, RefGTA. The scale of RefGTA is presented in Table~\ref{table:annotation_volumes}.
The resolution of the image is 1920$\times$1080. The mean length of annotated sentences is 10.06. We compared the saliency of the target using saliency model proposed by Itti \etal~\cite{Itti}, which is commonly used. First, we calculated a saliency map of a whole image, and we used the value in a bounding box of a target. As in Fig.~\ref{fig:salient_compare}, RefGTA contains more targets with low saliency as compared to RefCOCO. In this case, the relationships between the targets and salient context around them becomes more important.

\vspace{-3pt}
\section{Experiments}
\vspace{-2pt}
First, we explain the datasets used in our study. Next, we describe the results of comprehension, ranking, and generation evaluation in this order.
Finally, we evaluate the generated sentences by humans.

We refer to the state-of-the-art method for generation~\cite{Yu2017} as ``SLR.'' The SLR originally used VGGNet~\cite{Simonyan2015} as its image feature encoder. We also used
ResNet152 \cite{He2016} which achieved better performance on image classification. We compared re-implemented SLR and our model that we refer as ``re-SLR'', and ``our SR'' respectively. We set re-SLR with ResNet as a baseline. Our SLR implies our SR with the re-implemented listener.
 
\vspace{-2pt}
\subsection{Datasets}
\vspace{-2pt}
We conducted experiments on both existing datasets (RefCOCO, RefCOCO+~\cite{Yu2016} and RefCOCOg~\cite{Mao2016}) and our dataset (RefGTA). Our primarily purpose is the evaluation on RefGTA, whilst we used existing datasets to evaluate versatility of our method on various objects with real images. Yu \etal~\cite{Yu2017} collected more sentences for the test sets of RefCOCO and RefCOCO+; therefore, we used these for generation evaluation.

\vspace{-2pt}
\subsection{Comprehension Evaluation}
\vspace{-2pt}

\begin{table}[tbp]
\vspace{-6pt}
\scriptsize
\centering
\resizebox{\columnwidth}{!}{%
\begin{tabular}{| l | c | c | c | c | c |}
\hline
& \multicolumn{2}{c}{RefCOCO} & \multicolumn{2}{c|}{RefCOCO+} &RefCOCOg\\
\cline{2-6}
&\ \ Test A\ \  &\ \ Test B\ \   &\ \ Test A\ \  &\ Test B\ &\ val\\
\hline
SLR (ensemble) \cite{Yu2017}& 80.08\% & 81.73\% & 65.40\% & 60.73\% & 74.19\%\\
re-SLR (ensemble)			& 78.43\% & 81.33\% & 	64.57\% & 60.48\% & 70.95\%\\
\hline\hline
baseline: re-SLR (Listener) 		& \bf81.14\% & 80.80\% & \bf68.16\% & 59.69\% & 72.36\%\\
Our SR (Reinforcer) 	& 80.44\% & \bf81.04\% & 67.81\% & 58.97\% & \bf74.94\%\\
Our SLR (Listener) 	& 79.05\% & 80.31\% & 65.75\% & \bf62.18\% & 73.39\%\\
\hline
baseline: re-SLR (Speaker) 		& 80.70\% & 81.71\% & 68.91\% & 60.77\% & 72.55\%\\
Our SR (Speaker) 	& 82.45\% & \bf82.00\% & 72.07\% & \bf61.06\% & 70.35\%\\
Our SLR (Speaker) 	& \bf83.05\% & 81.84\% & \bf72.37\% & 59.13\% & \bf74.79\%\\
\hline
\end{tabular}}
\caption{Comprehension evaluation on RefCOCO, RefCOCO+ and RefCOCOg. () implies the model used. Ensemble implies to use both speaker and listener or reinforcer. Our speaker demonstrates comparable or better performance in most cases.}
\vspace{-10pt}
\label{table:comp_eval}
\end{table}

\begin{table}[tbp]
\scriptsize
\centering
\begin{tabular}{| l | c |}
\hline
& Test \\
\hline
baseline: re-SLR (Listener) &\bf86.05\% \\
Our SR (Reinforcer)			&85.25\% \\
Our SLR (Listener)			&84.04\% \\
\hline
baseline: re-SLR (Speaker) &84.84\% \\
Our SR (Speaker)			&88.41\% \\
Our SLR (Speaker)			&\bf88.60\% \\
\hline
baseline: re-SLR (ensemble) &86.55\% \\
Our SR (ensemble)		&89.16\% \\
Our SLR (ensemble)		&\bf89.54\% \\
\hline
\end{tabular}
\caption{Comprehension evaluation on RefGTA. Our speaker model exhibits high comprehension performance, and its ensembling exceeds that of re-SLR.}
\label{table:comp_our_eval}
\vspace{-12pt}
\end{table}

We compared comprehension performance of the speaker, listener, and reinforcer. Given a sentence $r$, each comprehension by reinforcer and speaker is calculated by $o^\ast=\argmax_{i}F(r, o_i)$ and $o^\ast=\argmax_{i}P(r|o_i)$, respectively. We used ground truth bounding boxes for all the objects. We only compared our method with the state-of-the-art model for generation~\cite{Yu2017} because our purpose is generation and we cannot compare methods for comprehension (e.g. ~\cite{Yu2018}) fairly.

\noindent{\textbf{Results on existing datasets:}} First, we demonstrate the comprehension performance on RefCOCO, RefCOCO+ and RefCOCOg in Table~\ref{table:comp_eval}. Although our speaker demonstrates comparable or better performance in most cases, we focus on the sentence generation, and the model with higher comprehension performance does not always generate better sentences.  Because both the listener and reinforcer used in~\cite{Yu2017} have a similar role as described in Sec.~\ref{sec:model}, we obtained similar results from our SR and our SLR.

\noindent{\textbf{Results on RefGTA:}} Next, we demonstrate the comprehension performance of the system on RefGTA in Table~\ref{table:comp_our_eval}. Although the listener's comprehension accuracy is better for re-SLR, our speaker's comprehension accuracy is higher than that of the re-SLR, and our model is best when ensembling a speaker and listener models. The accuracy on Table~\ref{table:comp_our_eval} is higher than the accuracies on Table~\ref{table:comp_eval} because we constructed large-scale dataset limiting targets to humans.

\vspace{-2pt}
\subsection{Ranking Evaluation on RefGTA}
\vspace{-2pt}
\begin{table}[tbp]
\vspace{-6pt}
\scriptsize
\centering
\begin{tabular}{| l | c |}
\hline
& Test \\
\hline
baseline: re-SLR (Reinforcer) &55.89\% \\
baseline: re-SLR (Speaker) &55.99\% \\
\hline
Our SR (Reinforcer) &55.46\% \\
Our SR (Speaker) &56.38\% \\
\hline
Our SR (Reinforcer) + rank loss &\bf57.55 \% \\
Our SR (Speaker)	+ rank loss	&56.64\%	\\
\hline
\end{tabular}
\caption{Accuracy of classifying ranked pairs. Ranking loss improved its performance.}
\label{table:under_our_eval}
\vspace{-5pt}
\end{table}
\begin{figure}[tbp]
\vspace{-5pt}
\begin{center}
  \includegraphics[width=0.87\linewidth]{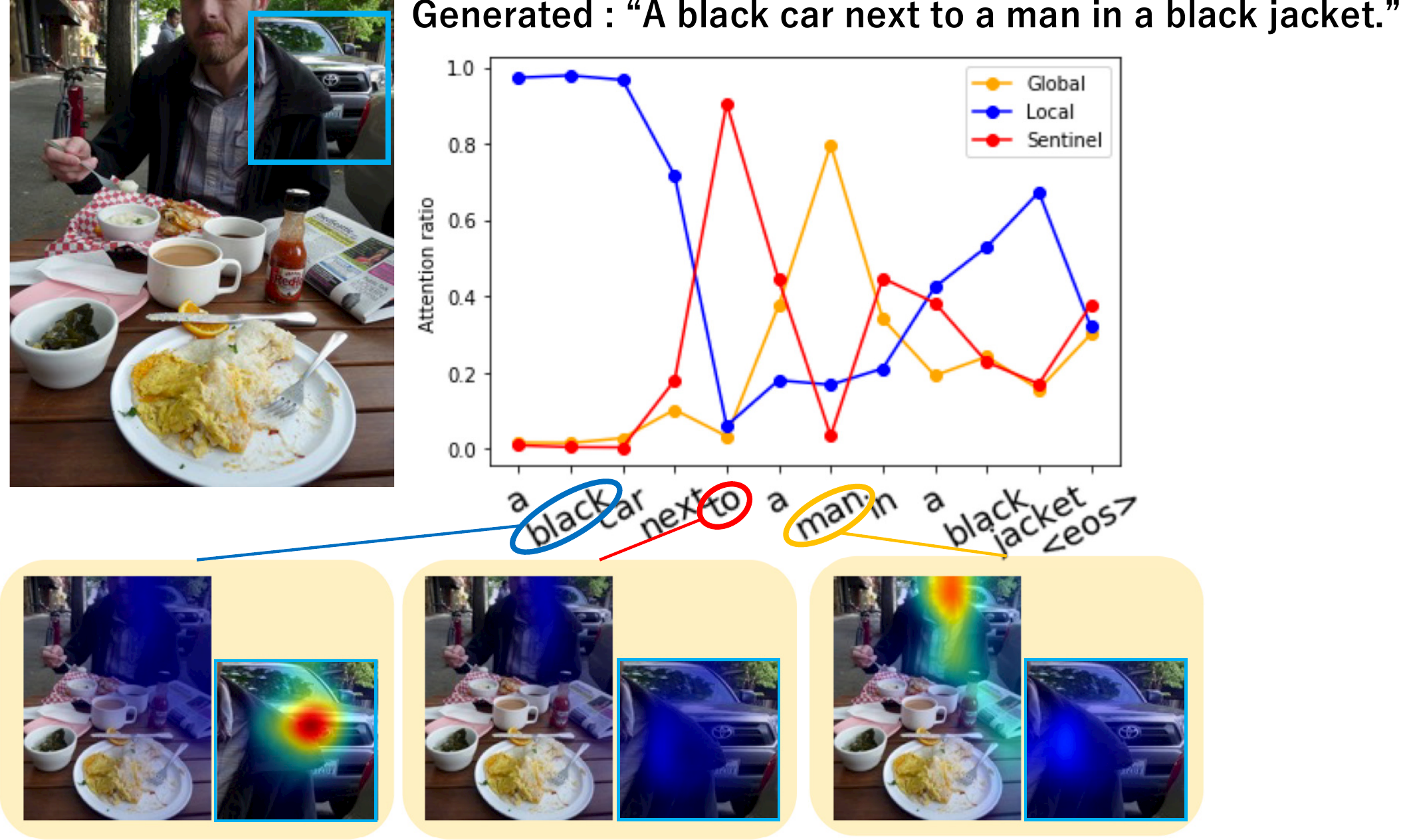}
  \caption{Generation example on RefCOCOg and each attention transition. Each of attention values corresponds to the sum of the softmax probability divided by local, global and sentinel in Eqn.~\ref{eqn:attn} respectively and their sum equals to one for each word.}
\label{fig:generation_attention}
\end{center}
\vspace{-24pt}
\end{figure}

We evaluated the ranking accuracy by classifying a given pair into two classes; whether the given two expressions are correctly ranked or not. First, we extracted the set of ranking pair as described in Sec.~\ref{sec:sent_ann}. The number of all pairs is 13,023. The results are shown in Table~\ref{table:under_our_eval}. ``Rank loss'' implies that we adopt the ranking loss for both speaker and reinforcer as we explain in Sec.~\ref{sec:ranking_opt}. Both of them improved the ranking performance by rank loss.
This implies that rank loss helps our model learning expressions comprehended by humans correctly and quickly.

\vspace{-2pt}
\subsection{Generation Evaluation}
\vspace{-2pt}

\begin{table*}[t!]
\vspace{-2pt}
\footnotesize
\begin{center}
\resizebox{1.7\columnwidth}{!}{%
\begin{tabular}{| l |l| c | c | c | c | c | c | c | c | c | c |}
\hline
&& \multicolumn{4}{c|}{RefCOCO} & \multicolumn{4}{c|}{RefCOCO+} & \multicolumn{2}{c|}{RefCOCOg}\\
\cline{3-12}
&features&  \multicolumn{2}{c|}{Test A} & \multicolumn{2}{c|}{Test B} &  \multicolumn{2}{c|}{Test A} & \multicolumn{2}{c|}{Test B} & \multicolumn{2}{c|}{val}\\
\cline{3-12}
&& Meteor & CIDEr & Meteor & CIDEr & Meteor & CIDEr & Meteor & CIDEr & Meteor & CIDEr \\
\hline\hline
SLR \cite{Yu2017}&VGGNet& {0.268} & {0.697} & {0.329} & {1.323} & {0.204} & {0.494} & 0.202 & {0.709} & 0.154 & {0.592}\\ 
SLR+rerank \cite{Yu2017}&VGGNet& {0.296} & {0.775} & {0.340} & {1.320} & {0.213} & {0.520} & 0.215 & {0.735} & 0.159 & {0.662}\\ 
\hline\hline
re-SLR  &VGGNet& 0.279 & 0.729 & 0.334 & 1.315 & 0.201 & 0.491 & 0.211 & 0.757 & 0.146 & 0.679\\
re-SLR+rerank   &VGGNet& 0.278 & 0.717 & 0.332 & 1.262 & 0.198 & 0.476 & 0.206 & 0.721 & 0.150 & 0.676\\
\hline
baseline: re-SLR	&ResNet& 0.296 & 0.804 & 0.341 & 1.358 & 0.220 & 0.579 & 0.221 & 0.798 & 0.153 & 0.742\\
\hline\hline
Our Speaker only &ResNet& 0.301 & \bf0.866 & 0.341 & \bf1.389 & \bf0.243 & \bf0.672 & 0.222 &\bf0.831 & 0.163 & 0.746\\
\hline
Our SR (w/o local attention) 	&ResNet& 0.289 & 0.760 & 0.328 & 1.278 & 0.214 & 0.542 & 0.210 & 0.753 & 0.156 & 0.666\\
Our SR (w/o global attention) 	&ResNet& 0.307 & 0.845 & 0.335 & 1.331 & 0.237 & 0.654 & 0.220 & 0.822 & 0.163 & 0.714\\
Our SR (w/o sentinel attention) 	&ResNet& 0.303 & 0.851 & 0.340 & 1.358 & 0.238 & 0.663 & 0.219 & 0.819 & 0.164 & 0.746\\
\hline
Our SR 	&ResNet& 0.307 & 0.865 & 0.343 & 1.381 & 0.242 & 0.671 & 0.220 & 0.812 & 0.164 & 0.738\\
Our SR+rerank 	&ResNet& 0.310 & 0.842 & \bf0.348 & 1.356 & 0.241 & 0.656 & 0.219 & 0.782 & 0.167 & 0.773\\
\hline
Our SLR &ResNet& 0.310 & 0.859 & 0.342 & 1.375 &0.241 &0.663 & 0.225 & 0.812 & 0.164 & 0.763\\
Our SLR+rerank&ResNet& \bf0.313 & 0.837 & 0.341 & 1.329 &0.242 &0.664 & \bf0.228 & 0.787 & \bf0.170 & \bf0.777\\
\hline
\end{tabular}
}
\end{center}
\vspace{-5pt}
\caption{Generation results using automatic evaluation. We used the test set of RefCOCO and RefCOCO+ extended by Yu \etal~\cite{Yu2017}. While the generation qualities are high by the speaker only in RefCOCO and RefCOCO+, rerank improves the performance in RefCOCOg.}
\label{table:metric}
\vspace{-12pt}
\end{table*}

\noindent{\textbf{Qualitative results on existing datasets:}} Generated sentence example on RefCOCOg is shown in Fig.~\ref{fig:generation_attention}. While the value of local attention is high when explaining the target car, the value of global attention becomes high when mentioning objects outside of the target. When switching from local attention to global attention, the value of sentinel attention that holds the sentence context becomes higher.

\noindent{\textbf{Quantitative results on existing datasets:}} Next, we discuss the quantitative evaluation based on the automatic evaluation metric, CIDEr~\cite{Vedantam2015} and Meteor~\cite{Lavie2007}. Because ground-truth sentences are referring expressions, we can evaluate them to some extent.
Our re-implemented rerank did not improve the generation performance although Yu \etal~\cite{Yu2017} reported that reranking improves performance.
In RefCOCO and RefCOCO+, the generation qualities are high by the speaker only. Meanwhile, in RefCOCOg, rerank helped to improve the performance. This is because while the model should generate one phrase in RefCOCO and RefCOCO+, the model should generate a full sentence in RefCOCOg and has to solve more complex problems including satisfying language structures.

\begin{figure*}[t!]
\centering
\vspace{-5pt}
  \includegraphics[width=\linewidth]{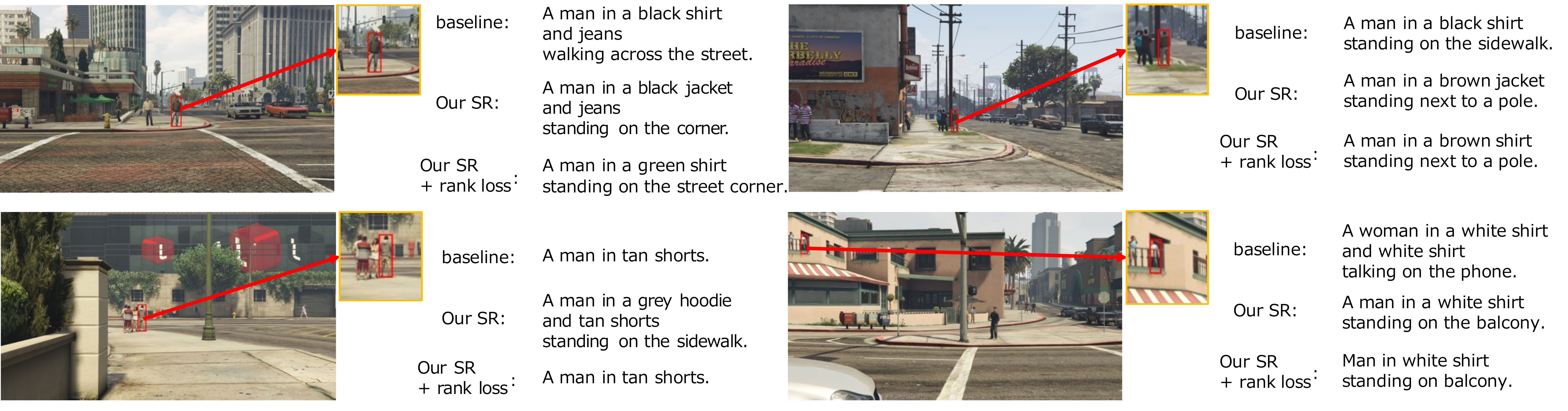}
  \caption{Comparison of generated sentences by each method on RefGTA. Rank loss implies to be trained with ranking.}
\vspace{-4pt}
\label{fig:gen_out_attention_refgta2}
\end{figure*}

\begin{figure*}[t!]
\centering
\vspace{-5pt}
  \includegraphics[width=0.95\linewidth]{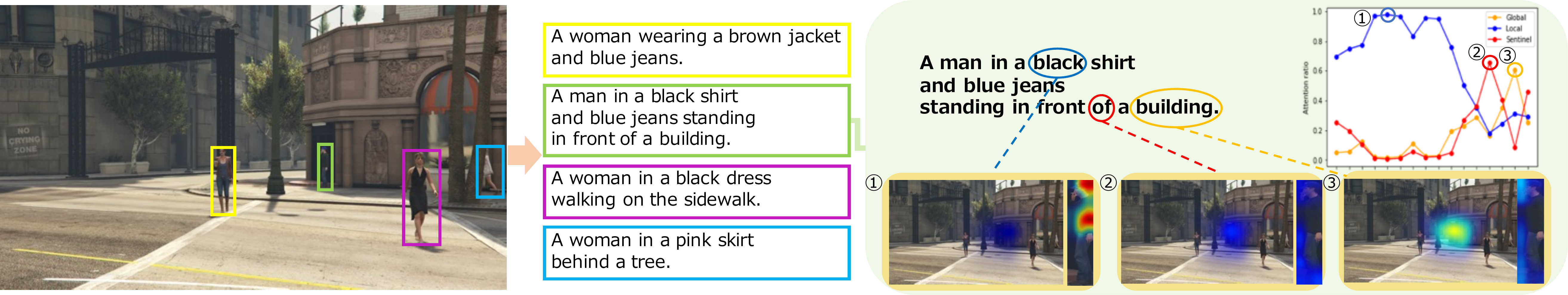}
  \caption{Generation example on RefGTA and each attention transition. Each sentence is generated from an object of the same color.}
\label{fig:gen_out_attention_refgta1}
\vspace{-13pt}
\end{figure*}

\noindent{\textbf{Qualitative results on RefGTA:}} Next, we demonstrate the generated sentence examples on RefGTA in Fig.~\ref{fig:gen_out_attention_refgta2}.
While the baseline method (re-SLR) demonstrates lower capability in capturing the outside of the target than our method, our method can generate sentences that can identify the target easier especially in the right-side examples. As shown in the left-bottom example, while the baseline method generates a brief and sufficient description, our SR+rank loss also generates the same one. Attention visualization is shown in Fig.~\ref{fig:gen_out_attention_refgta1}.
While the local attention value is high when describing the target, the global attention value is high when mentioning ``building,'' which is outside of the target.

\noindent{\textbf{Quantitative results on RefGTA:}} Finally, we demonstrate the quantitative evaluation on RefGTA.
In our study, the ideal metric should assign a high score to a sentence that can be easily comprehended by humans correctly and quickly. While CIDEr calculates the average similarity between a generated sentence from an object $o_i$ and ground-truth sentences $\{r_{i1},\cdots,r_{im}\}$; we define the ranking-weighted CIDEr (R-CIDEr) which utilizes weighted similarity scores between them by the inverse of their rank. The weight of the sentence $r_{ij}$ is calculated as $w(r_{ij})=\left( rank(r_{ij})\sum_j rank(r_{ij})^{-1} \right)^{-1}$. This metric assigns a high score to sentences where a human identified the referred objects correctly and quickly. In Table~\ref{table:gen_our_eval}, R1-CIDEr implies using ranking by humans' comprehension accuracy and time required, and R2-CIDEr implies using ranking by only humans' comprehension accuracy. In particular, R1-CIDEr that we optimized is improved by the ranking loss. Rerank was not applicable in RefGTA.

\begin{table}[tbp]
\scriptsize
\centering
\resizebox{\columnwidth}{!}{%
\begin{tabular}{| l | c | c| c| c|}
\hline
&  \multicolumn{4}{c|}{Test} \\
\cline{2-5}
& Meteor & CIDEr & R1-CIDEr & R2-CIDEr\\
\hline\hline
baseline: re-SLR		&0.263&0.966&0.994&0.976\\
\hline
Our Speaker only			&0.278 &1.014 &1.038&1.025\\
\hline
Our SR (w/o local attention)		&0.208 &0.557 &0.570&0.561\\
Our SR (w/o global attention)	&0.276 &1.036 &1.065&1.047\\
Our SR (w/o sentinel attention)	&0.278 &1.022 &1.049&1.033\\
Our SR			&\bf0.279 &1.036 &1.065&1.048\\
Our SR+rank loss	&0.277 &\bf1.047 &\bf1.078&\bf1.059\\
\hline
Our SLR			&0.278 &1.030 &1.054&1.041 \\
\hline
\end{tabular}
}
\caption{Generation evaluation on RefGTA. Without rank loss, Our SR with all modules is the best. Furthermore, ranking improved performance.}
\label{table:gen_our_eval}
\end{table}

\subsection{Human Evaluation on RefGTA}

\begin{table}[tbp]
\begin{center}
\vspace{-8pt}
\resizebox{\columnwidth}{!}{
\begin{tabular}{|l|c|c|c||c|c|}
\hline
 & Unspecified color & Black wearing & White wearing & All & All (selected)\\
\hline
baseline  & 69.52\% & 60.96\% & 71.09\% & 67.48\% &73.17\% \\
\hline
Our SR  & \bf75.67\% & 64.61\% & \bf72.93\% & 71.52\% &75.83\% \\
\hline
Our SR+rank loss & 74.89\% & \bf68.76\% & 72.30\% & \bf72.25\%   &\bf76.94\%\\
\hline
\end{tabular}}
\caption{Left three columns: the evaluation of humans' comprehension accuracy when divided by clothing types as seen in Fig.~\ref{fig:image_collect}, All: The rate for which annotators were able to select the correct target, All (selected): The accuracy ignoring ``impossible to identify'' choices.}
\label{table:gen_human_acc}
\end{center}
\vspace{-20pt}
\end{table}

\noindent{\textbf{Human comprehension evaluation:}} First, we evaluated human comprehension for the generated sentences by each method. We used 600 targets extracted randomly from the test data, and requested 10 AMT workers to identify the referred persons while measuring the time. If no referred target exists, we allow them to check a box, ``impossible to identify.'' The results including clothing type evaluations are shown in Table~\ref{table:gen_human_acc}. Our model 
outperformed the baseline method, and the rank loss improved the performance in black wearing case. Our SR+rank loss was the best for the overall performance.

\noindent{\textbf{Time evaluation:}} Next, we evaluated whether our method improved performance based on the time required by humans to locate referred objects. We evaluated as follows: first, all sentences are ranked by humans' comprehension accuracy; subsequently, sentences that are comprehended correctly by all workers (i.e., comprehension accuracy is 100\%) are ranked by the average time; finally, for the remaining sentences, we calculated the ratio of the number of instances that are ranked first in each method (if there are 2 or 3 instances ranked first, the number is counted as 1/2, and 1/3 respectively.) 
The obtained results (see Table~\ref{table:human_time}) show that rank loss improved not only human comprehension accuracy but also the time.

\begin{table}[tbp]
\begin{center}
\resizebox{0.9\columnwidth}{!}{
\begin{tabular}{|l|c|c|}
\hline
 &Accuracy only& Accuracy and time\\
\hline
  baseline & 30.08\% & 30.86\% \\
  Our SR   & 34.08\% & 33.28\% \\
  Our SR+rank loss & \bf35.83\% & \bf35.86\% \\
\hline
  difference between proposed methods & 1.75\% & 2.58\% \\
\hline
\end{tabular}}
\caption{Comparison of our generated sentences in terms of humans' comprehension accuracy and the time required to locate the referred objects. For all methods, the sum of accuracies is 100\%. When including time in the comparison, the difference between the proposed methods increases. This shows the efficacy of using rank loss.}
\label{table:human_time}
\end{center}
\vspace{-17pt}
\end{table}

\begin{figure}[tbp]
\begin{center}
  \includegraphics[width=0.9\linewidth]{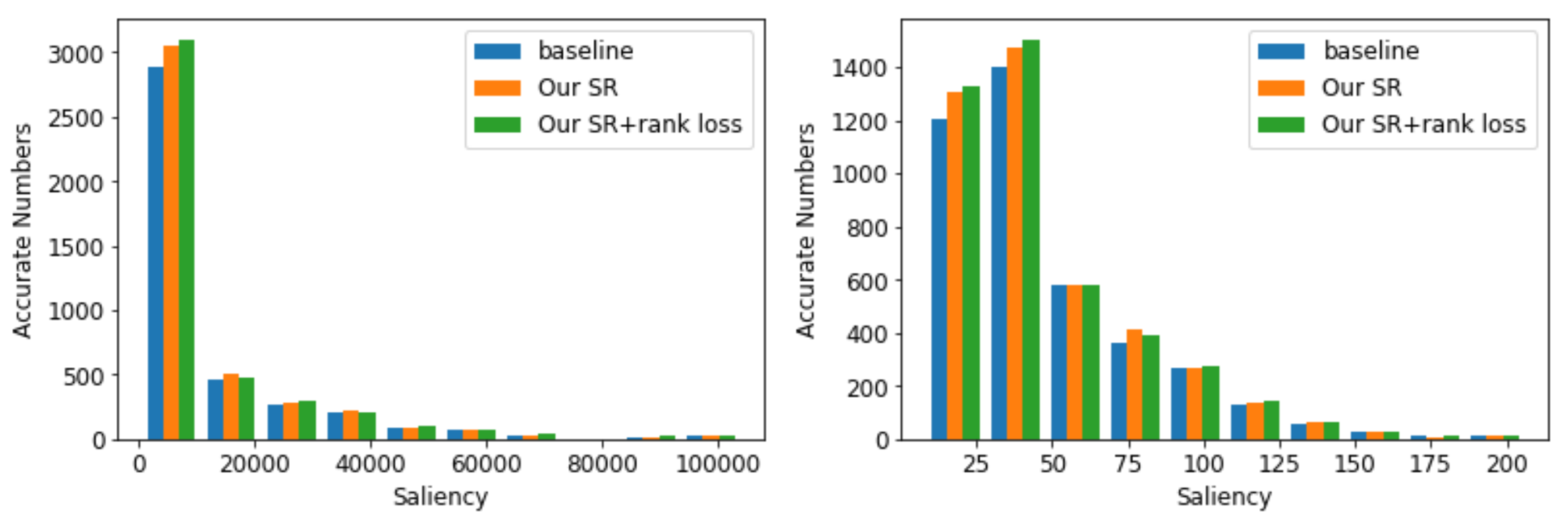}
\vspace{-4pt}
  \caption{Relationship between the number of people who answered correctly and saliency calculated as Fig.~\ref{fig:salient_compare}.}
\vspace{-23pt}
\label{fig:human_sal_acc}
\end{center}
\end{figure}

\noindent{\textbf{Human comprehension evaluation considering saliency:}} Finally, we evaluated our method when the saliency of the target changes. We evaluated the relationship between humans' comprehension accuracy and targets' saliency. We calculated saliency as described in Sec.~\ref{sec:stat_info}. We present the results in Fig.~\ref{fig:human_sal_acc}. As shown, our model performs better on the lower saliency area because mentioning salient contexts around the targets helped humans to comprehend them. The difference between the methods becomes smaller as the saliency becomes higher.

\section{Conclusions}
We herein focused on generating referring expressions that allowed for humans to identify referred objects correctly and quickly. We proposed a model that could utilize relationships between targets and contexts around them to generate better sentences even when the compositions of the images were complex, and the targets were not sufficiently salient. We also proposed a method to optimize referring expressions that are easy for target identifications with additional annotations. To evaluate our system, we constructed a new dataset, RefGTA. We demonstrated that our method improved referring expression generation not only on the existing automatic evaluation metric, but also on the newly proposed automatic evaluation metric and human evaluation.

\section*{Acknowledgment}
\addcontentsline{toc}{section}{Acknowledgment}
This work was supported by JST ACT-I Grant Number JPMJPR17U5, partially supported by JSPS KAKENHI Grant Number JP17H06100 and partially supported by JST CREST Grant Number JPMJCR1403, Japan.
We would like to thank Atsushi Kanehira, Hiroaki Yamane and James Borg for helpful discussions.

{\small
\bibliographystyle{ieee_fullname}
\bibliography{egbib}
}

\renewcommand{\thesection}{\Alph{section}}
\setcounter{section}{0}

\clearpage

\twocolumn[
\maketitle\vspace*{54pt}
\begin{center}{\Large \bf Supplementary Material}\vspace*{54pt}\end{center}
]

\section{Our task from the viewpoint of Gricean Maxims}

Gricean Maxims~\cite{Grice1975}, which is advocated as a collaborative principle for effective conversation between a speaker and a listener, has often been discussed in referring expression generation~\cite{Dale1995,Golland2010,Yu2016}. 
Gricean Maxims has four aspects: quality, quantity, relation and manner. When the targets are salient like on RefCOCO, the evaluation of the comprehension accuracy is enough to satisfy Gricean Maxims. However, when the composition of the image becomes complex like Fig.~1, the comprehension time which relates to quantity and manner is also needed for sentence evaluations to satisfy Gricean Maxims.

\section{The novelty of our ``context''}
``Context'' in our paper refers to the visual context of the target, such as nearby objects or features and also the context during generation of sentences where context here refers to previously generated words in the sentence. The visual context of the target allows us to identify its global location whilst also distinguishing from other targets. We back propagated the loss for generating expressions uniquely referring to the target back through to global, local and sentinel attention in Fig. 2. Our model can generate sentences by selecting important information for identification from inside and outside the target bounding box. ``Context'' plays an important role especially in such cases where the target is less salient or the target is hard to refer to by just mentioning its attributes. 

Existing referring expression generation research~\cite{Yu2016} also uses ``context.'' This research aims to distinguish the target from others however does not attempt to inform us of the location and does not utilize the relationship of the target to nearby objects or features which are not the same class as the target in their ``context.''

\section{Dataset comparison}

\subsection{Size}

\begin{table}[t!]
\begin{center}
\resizebox{\columnwidth}{!}{%
\begin{tabular}{|l|c|c|c|c|c|}
\hline
 & RefCOCO & RefCOCO+ & RefCOCOg & \cite{Vasudevan2018} (Video) & RefGTA \\
\hline\hline
\# of images  & 19,994 & 19,992 & 25,799 & 4,818 &28,750\\
\# of created instances & 50,000 & 49,856 & 49,822 & 29,901 & 78,272\\
\# of referring expressions & 169,806 & 166,403 & 95,010& 30,320 & 213,175\\
\hline
\end{tabular}}
\caption{Statistics of annotations on existing datasets and our dataset (RefGTA). RefGTA contains more images, instances and referring expressions than the other datasets.}
\label{table:all_annotation_volumes}
\end{center}
\end{table}

The comparison of the size in existing datasets (RefCOCO, refCOCO+~\cite{Yu2016}, RefCOCOg~\cite{Mao2016} and ~\cite{Vasudevan2018}) and RefGTA is shown in Table~\ref{table:all_annotation_volumes}.
RefGTA contains more images, instances and referring expressions than existing datasets.
The number of instances and the number of referring expressions are almost the same in \cite{Vasudevan2018} because the purpose of \cite{Vasudevan2018} is comprehension and does not need multiple sentences for automatic evaluation like generation.

\subsection{The reason why RefGTA tends to have more targets with lower saliency than RefCOCO}
\begin{figure}[t!]
\begin{center}
  \includegraphics[width=0.85\linewidth]{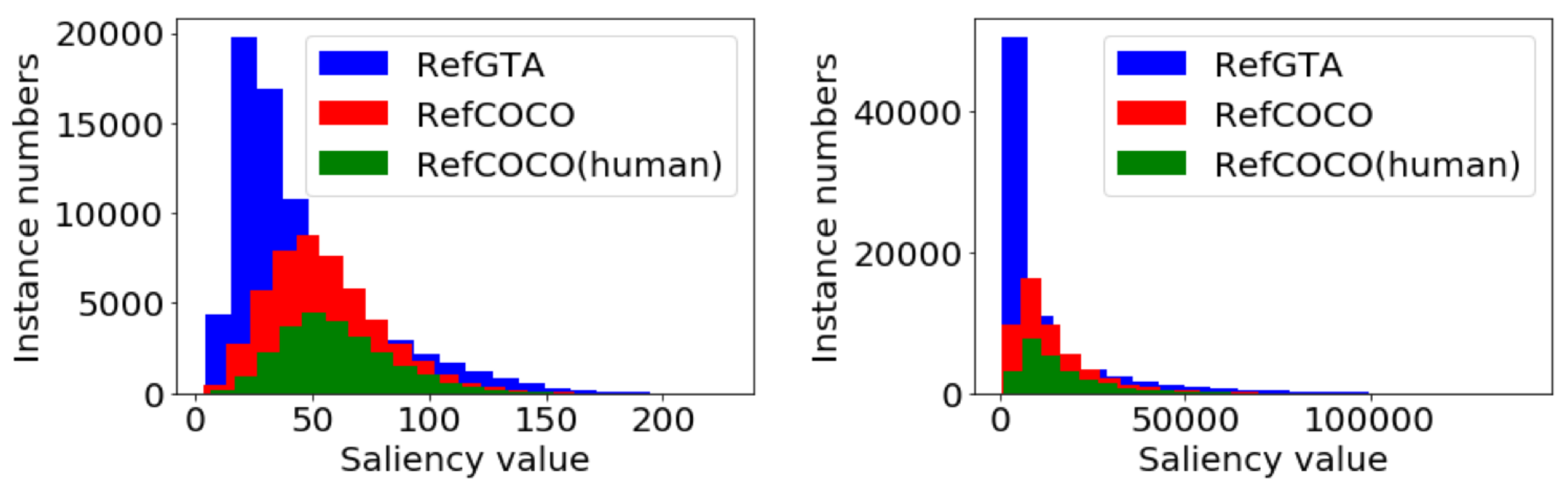}
  \caption{Targets' saliency of RefGTA, RefCOCO and RefCOCO (human) calculated as Fig.~5. As the saliency becomes higher, the ratio of human instances becomes larger in RefCOCO.}
\label{fig:sal}
\end{center}
\end{figure}

As shown in Fig.~\ref{fig:sal}, even if we limit targets to humans in RefCOCO, there are few instances with low saliency. Images captured automatically are different from images taken by a person as they do not have subjects and tend to have miscellaneous information.

\subsection{Word distribution}

\begin{figure*}[t!]
\begin{center}
  \includegraphics[width=\linewidth]{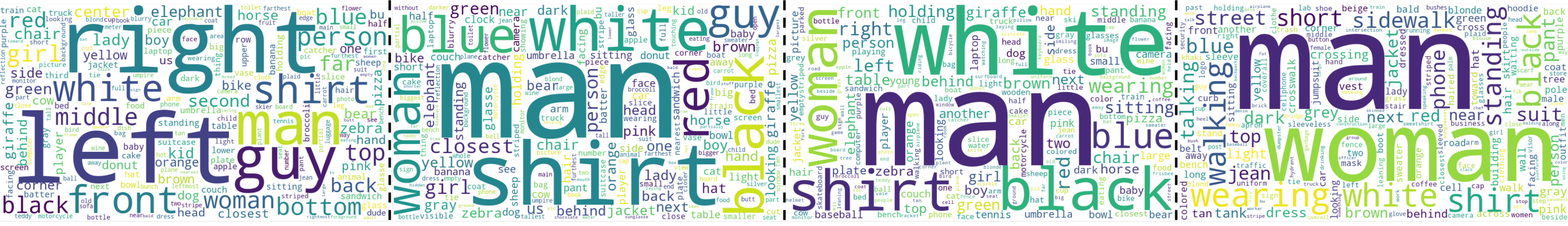}
  \caption{Word clouds (from left to right: RefCOCO, RefCOCO+, RefCOCOg, RefGTA)}
\label{fig:wcloud}
\end{center}
\end{figure*}

The word clouds on RefCOCO, RefCOCO+, RefCOCOg and RefGTA are shown in Fig.~\ref{fig:wcloud}.

\section{Detailed results}

\subsection{Vocabulary in the generated sentences}
\begin{table}[t!]
\begin{center}
\resizebox{0.8\columnwidth}{!}{
\begin{tabular}{|l|c|c|c|}
\hline
 & baseline & Our SR & Our SR + rank loss \\
\hline
\# of vocabularies & 134 & 172 & 176 \\
\hline
\end{tabular}}
\caption{The number of vocabulary in the sentences generated by each method in the test set on RefGTA.}
\vspace{-5pt}
\label{table:ref_vocab}
\end{center}
\end{table}

The number of vocabulary in the sentences generated by each method on RefGTA is shown in Table~\ref{table:ref_vocab}.
Both of the sentences generated by our methods contain more vocabularies to represent the targets' surroundings (such as ``beach'', ``bridge'', ``bus'', ``palm'', ``stairs'', ``store'' , ``pillar'', ``plant'' , ``railing'' and ``truck'') than the sentences generated by the baseline method.

\subsection{Generated sentence length}
\begin{table}[t!]
\begin{center}
\resizebox{\columnwidth}{!}{
\begin{tabular}{|l|c|c|c|c|}
\hline
 & RefCOCO (test) & RefCOCO+ (test) &RefCOCOg (val) &RefGTA (test)\\
\hline
baseline  & 2.68 & 2.51 & 6.50 & 9.46\\
\hline
Our SR  & 2.93 & 2.63 & 7.27 & 10.18\\
\hline
Our SR + rank loss  & - & - & - & 9.82\\
\hline
\hline
Ground-Truth  & 3.71 & 3.58 & 8.48 & 10.04\\
\hline
\end{tabular}}
\caption{The average lengths of generated and ground-truth sentences. Our SR generated longer sentences than the baseline method in all datasets.}
\label{table:data_sent}
\end{center}
\end{table}

The mean lengths of generated and ground-truth sentences (i.e., the number of words in a description) are shown in Table~\ref{table:data_sent}.

Our SR generated longer sentences than the baseline method. Considering that our method improved the automatic evaluation metrics as shown in Table 5 and Table 6, this indicates that our SR has the ability to describe in more detail than the baseline method. However, our SR generated shorter sentences than ground-truth in existing datasets, indicating that there is still a need for improvement in the capability of describing the targets. 

On the other hand, the comprehension difficulty of GT varies on RefGTA unlike RefCOCO/+/g. Our SR generated sentences as long as ground-truth in RefGTA because RefGTA is a large-scale dataset limiting targets to humans. This enabled our model to focus on easy-to-understand referring expression generation, and our SR+rank loss learned the concise sentences that are relatively easy to be comprehended.

The mean lengths of generated sentences on RefGTA grouped by GT lengths are also shown in Table~\ref{table:ref}. Our SR+rank loss generated longer sentences than the baseline while shorter than our SR in each group.

\begin{table}[tbp]
\begin{center}
\resizebox{0.6\columnwidth}{!}{
\begin{tabular}{|l|c|c|c|}
\hline
 & short GT & middle GT & long GT \\
\hline
baseline  & 9.26 & 9.48 &9.62 \\
\hline
Our SR  & 9.87 & 10.28 & 10.39 \\
\hline
Our SR+rank loss & 9.54 & 9.90 & 10.02 \\
\hline
GT  & 7.21 & 9.81 & 13.17 \\
\hline
\end{tabular}}
\caption{
The average sentence lengths when instances are divided equally into three parts by calculating the average lengths of GT.}
\label{table:ref}
\end{center}
\end{table}

\subsection{Qualitative results}
\begin{figure*}[tbp]
\centering
  \includegraphics[width=\linewidth]{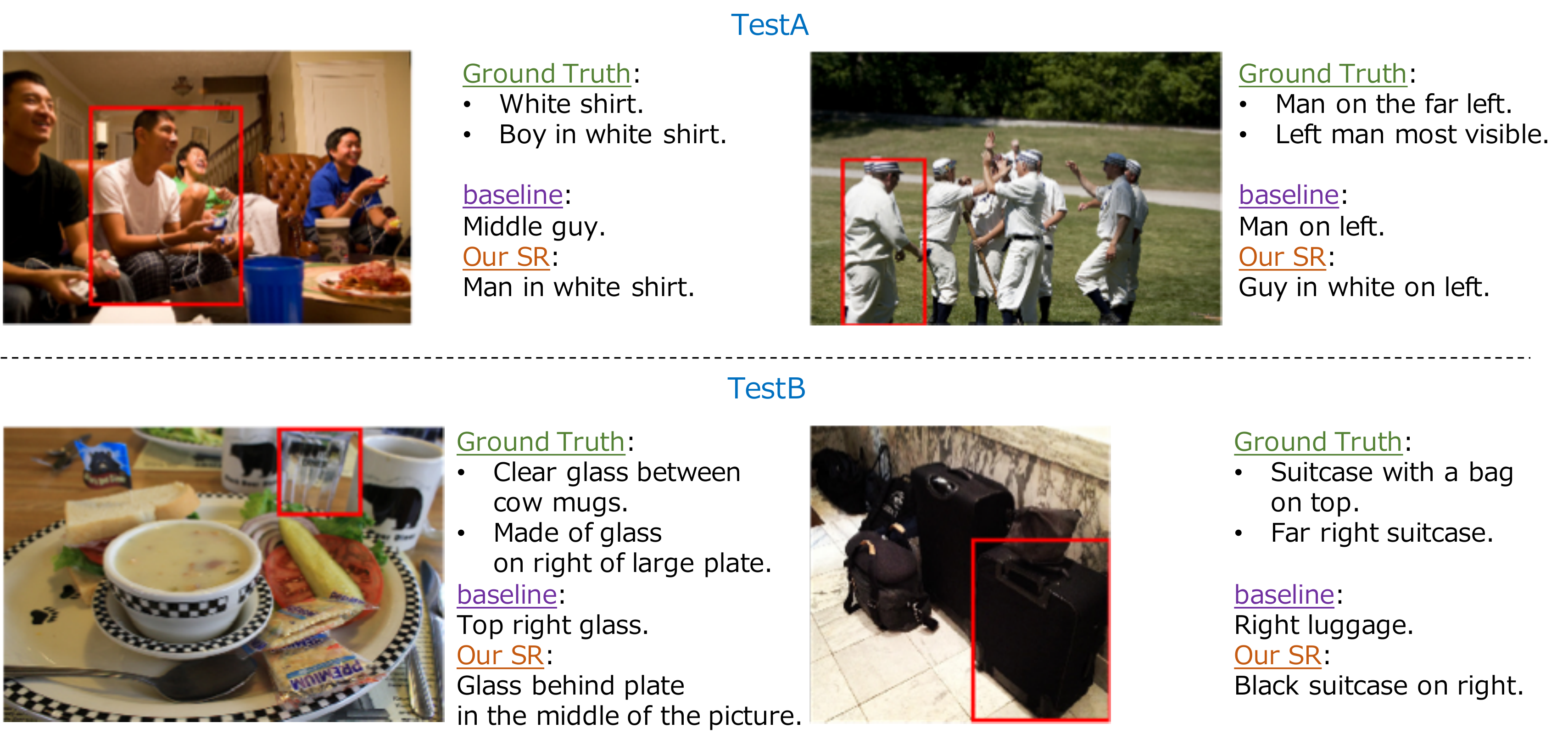}
  \caption{Generation example on RefCOCO by different methods with two ground-truth captions.}
\label{fig:co_gt}
\end{figure*}

\begin{figure*}[tbp]
\centering
  \includegraphics[width=\linewidth]{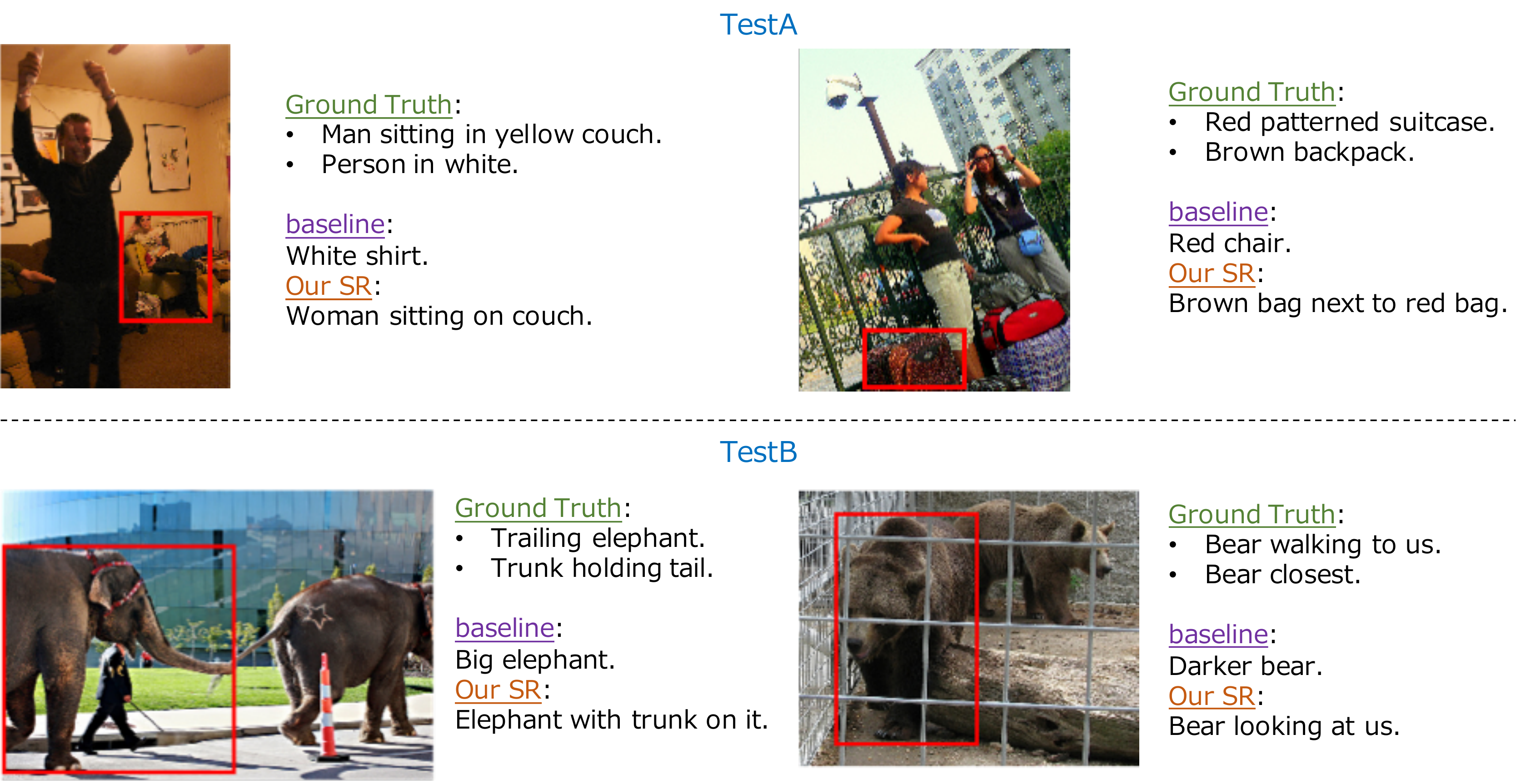}
  \caption{Generation example on RefCOCO+ by different methods with two ground-truth captions.}
\label{fig:co+_gt}
\end{figure*}

\begin{figure*}[tbp]
\centering
  \includegraphics[width=\linewidth]{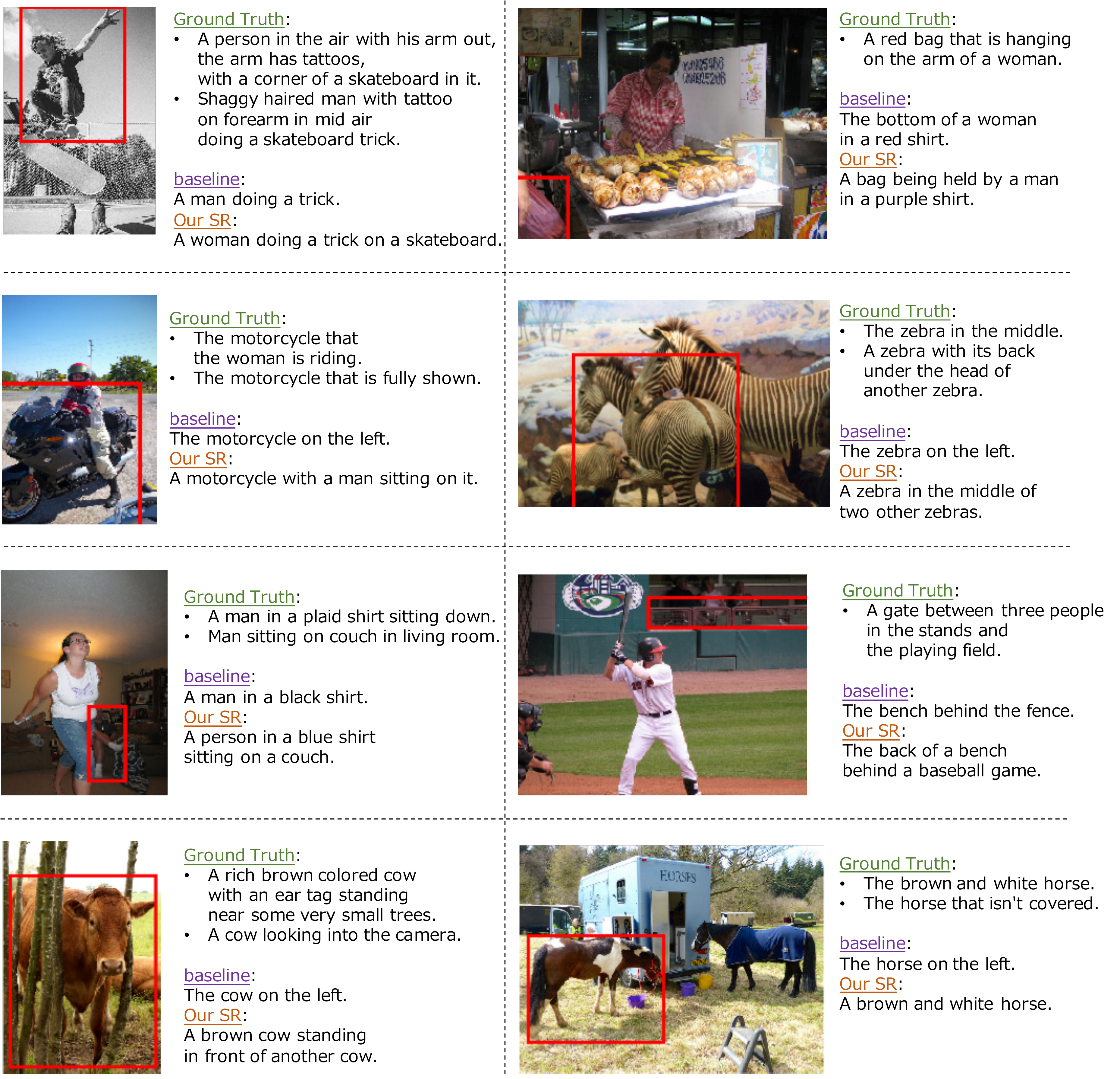}
  \caption{Generation example on RefCOCOg by different methods with one or two ground-truth captions. }
\label{fig:cog_gt}
\end{figure*}

\begin{figure*}[tbp]
\centering
  \includegraphics[width=0.95\linewidth]{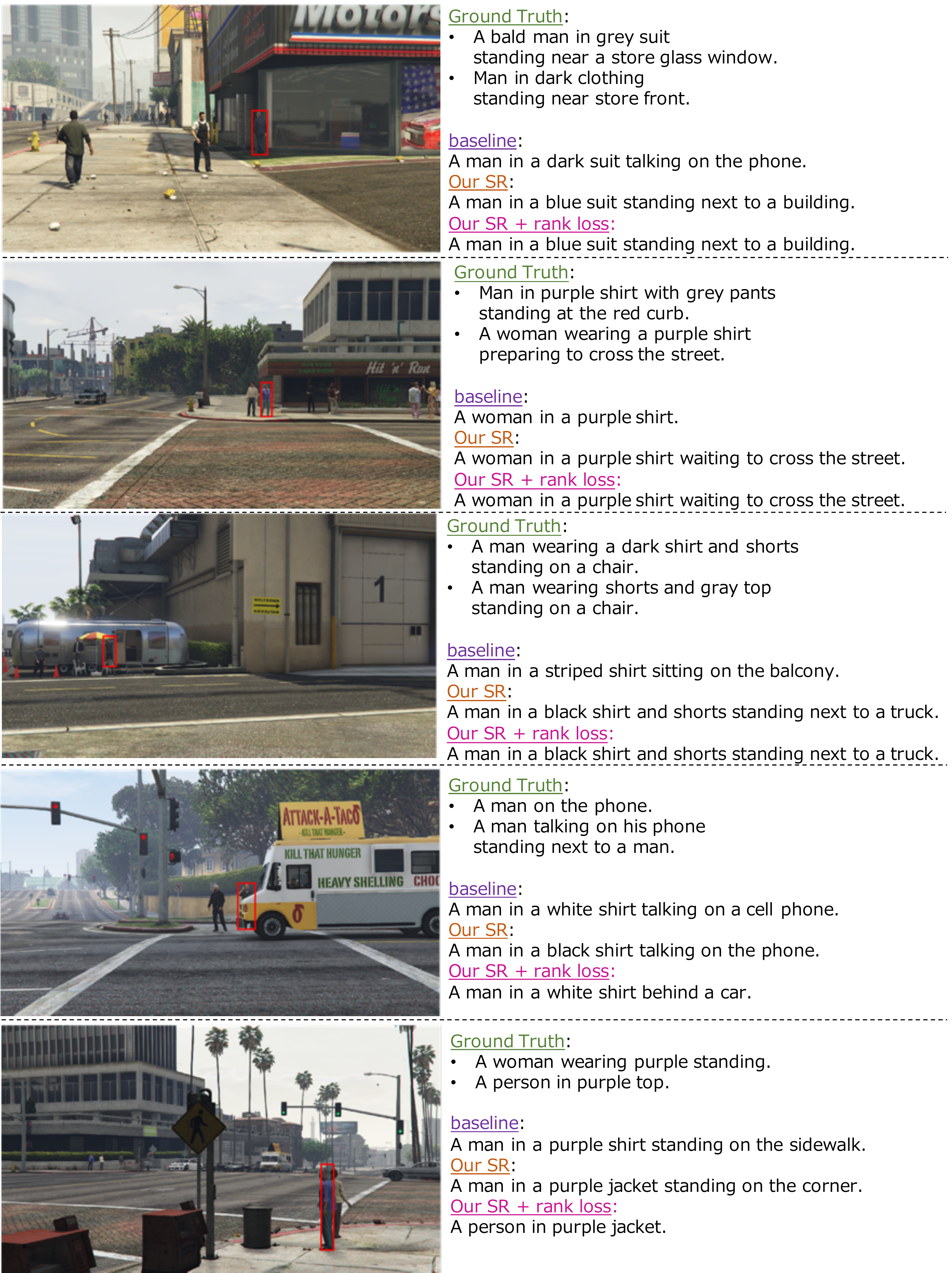}
  \caption{Generation example on RefGTA by different methods with two ground-truth captions.}
\label{fig:gta_gt}
\end{figure*}

In this paper, we showed qualitative results on existing datasets (RefCOCO, RefCOCO+ and RefCOCOg) and our dataset (RefGTA). Here we show more results in Fig.~\ref{fig:co_gt}, Fig.~\ref{fig:co+_gt}, Fig.~\ref{fig:cog_gt} and Fig.~\ref{fig:gta_gt}.

\end{document}